\documentclass[11pt]{article}

\usepackage{acl}
\usepackage{times}
\usepackage{latexsym}
\usepackage{listings}
\usepackage{amsmath}
\usepackage{multirow}
\usepackage[T1]{fontenc}
\usepackage[utf8]{inputenc}
\usepackage{microtype}
\usepackage{inconsolata}
\usepackage{graphicx}
\usepackage{booktabs}

\title{From Guidelines to Guarantees: A Graph-Based Evaluation Harness for Domain-Specific Evaluation of LLMs}

\author{
Jessica M. Lundin \\
Usman Nasir Nakakana \\
Guillaume Chabot-Couture \\
Gates Foundation \\
}

\begin{document}
\maketitle

\begin{abstract}
Rigorous evaluation of domain-specific language models requires benchmarks that are comprehensive, contamination-resistant, and maintainable. Static, manually curated datasets do not satisfy these properties. We present a graph-based evaluation harness that transforms structured clinical guidelines into a queryable knowledge graph and dynamically instantiates evaluation queries via graph traversal. The framework provides three guarantees: (1) complete coverage of guideline relationships; (2) surface-form contamination resistance through combinatorial variation; and (3) validity inherited from expert-authored graph structure. Applied to the WHO IMCI guidelines, the harness generates clinically grounded multiple-choice questions spanning symptom recognition, treatment, severity classification, and follow-up care. Evaluation across five language models reveals systematic capability gaps. Models perform well on symptom recognition but show lower accuracy on treatment protocols and clinical management decisions. The framework supports continuous regeneration of evaluation data as guidelines evolve and generalizes to domains with structured decision logic. This provides a scalable foundation for evaluation infrastructure.
\end{abstract}

\paragraph*{Data and Code Availability}
The WHO IMCI handbook is publicly available \cite{WHO2014imci}. Our graph construction, question generation code, and generated question dataset are available at \url{https://github.com/jessicalundin/graph_testing_harness}.

\section{Introduction}

\subsection{The Evaluation Coverage Problem}
Rigorous evaluation of language models faces a critical challenge: the distribution gap between application-specific text and existing benchmark datasets. This gap encompasses both context (domain, localization, complexity) and coverage (tasks, content). Current medical benchmarks rely on human curation, which is resource-intensive and results in incomplete coverage of specific medical guidelines.

MCQA benchmark datasets serve dual purposes: training new models and evaluating across models. The test split has widespread utility as a yardstick for comparison across models. While vignettes and multi-turn conversation with evaluation rubrics \cite{tu2024conversational,nori2025sequential,arora2025healthbench} more closely resemble real-world scenarios, MCQA remains an important evaluation format because it is less ambiguous, easy to grade, and scalable.

Despite advances in model architectures and training paradigms, MCQA benchmarks remain central for both evaluation and post-training. In health-domain models, supervised finetuning continues to be useful. Within alignment, MCQA also provides naturally ranked outputs for methods such as GRPO, where correct answers serve as high-reward samples and incorrect options serve as progressively lower-reward samples without requiring expensive human ranking.

WHO guidelines are an appropriate use case for this setting because there is substantial need for AI systems that support scarce healthcare workers in low- and middle-income countries (LMICs). These guidelines are often country-specific, which makes custom evaluation necessary for accurate measurement of model performance.

\subsection{Limitations of Existing Medical Benchmarks}
Medical benchmarks exist in multiple languages, and rely on questions from licensing exams, textbooks, journals, and crowdsourcing \cite{jin2021disease,pal2022medmcqa,vilares2019head,labrak2022frenchmedmcqa,kasai2023evaluating,jin2019pubmedqa,zhang2017chinese,olatunji2024afrimedqa,hendrycks2021measuring,alonso2024medexpqa}. Synthetic medical QA datasets employ diverse generation strategies: template-based approaches as in emrQA \cite{pampari2018emrqa} and RadQA \cite{soni2022radqa}, generation using ontology concepts \cite{dong2023ontology}, and LLM-based generation for hallucination detection \cite{pal2023medhalt}.

Existing MCQA benchmarks differ from our approach in three important ways. First, they rely on static question sets drawn from licensing exams, textbooks, and crowdsourcing, which are vulnerable to contamination as models are trained on increasingly broad corpora. Second, they provide aggregate scores that obscure performance on specific clinical relationships: a model may score well overall while systematically failing on treatment protocols or follow-up schedules. Third, they do not provide coverage guarantees relative to any specific guideline, making it impossible to know which relationships have and have not been tested. Non-MCQA evaluation formats such as patient vignettes \cite{tu2024conversational} and multi-turn conversations with evaluation rubrics \cite{nori2025sequential,arora2025healthbench} 
more closely approximate real clinical reasoning but are expensive to construct, difficult to grade consistently, and cannot be regenerated as guidelines evolve. Graph-based MCQA occupies a complementary position: it provides the discrete gradability and scalability of MCQA with 
coverage guarantees and contamination resistance that static benchmarks lack, while serving as a structured precursor to higher-stakes evaluation in more realistic formats.

\subsection{Contributions}
Our main contributions are as follows:
\begin{enumerate}
    \item We introduce a graph-based evaluation harness that provides explicit guarantees of coverage, contamination resistance, and validity.
    \item We present a method for transforming structured clinical guidelines into a knowledge graph that supports systematic evaluation.
    \item We demonstrate dynamic evaluation through on-demand query instantiation rather than static datasets.
    \item We empirically show that this framework reveals systematic weaknesses in clinical reasoning that are not captured by aggregate benchmarks.
\end{enumerate}

\section{Method}

\subsection{Graph Construction from Clinical Guidelines}
We transform the WHO IMCI handbook \cite{WHO2014imci} into a directed graph structure. The handbook, an 80-page document containing flowcharts and checklists for childhood illness management, is parsed to extract medical entities and their relationships. The resulting graph contains 200+ nodes and 300+ edges spanning respiratory, gastrointestinal, nutritional, and infectious diseases.

The graph schema consists of five node types:
\begin{itemize}
    \item \textbf{Condition} (31 nodes): Medical conditions with age range attributes (0--2 months for young infants, 2--60 months for children)
    \item \textbf{Symptom} (79 nodes): Observable clinical indicators (e.g., ``fast breathing'', ``convulsions'')
    \item \textbf{Treatment} (84 nodes): Medical interventions (e.g., ``give oral Amoxicillin for 5 days'')
    \item \textbf{FollowUp} (15 nodes): Monitoring schedules (e.g., ``3 days'', ``7 days'')
    \item \textbf{Severity} (4 nodes): Triage classifications (severe, moderate, mild, none)
\end{itemize}

Four edge types connect these nodes:
\begin{itemize}
    \item \textbf{INDICATES}: Symptom $\rightarrow$ Condition
    \item \textbf{TREAT}: Condition $\rightarrow$ Treatment
    \item \textbf{FOLLOW}: Condition $\rightarrow$ FollowUp
    \item \textbf{TRIAGE}: Condition $\rightarrow$ Severity
\end{itemize}

Automated extraction via PDF parsers and LLMs failed to reliably capture the conditional logic embedded in IMCI flowcharts. Relationships expressed visually through color-coded triage paths and nested decision branches cannot be faithfully reconstructed as directed edges by current PDF and LLM pipelines. The knowledge graph was therefore manually curated by a co-author with over 
15 years of clinical practice, specialized pediatric training, and extensive experience implementing WHO IMCI guidelines in sub-Saharan Africa. Curation proceeded in three stages: (1) the clinical expert parsed each flowchart and checklist page to identify entity mentions and candidate relationships; (2) candidate edges were encoded in a structured schema and reviewed against the source document for completeness; and (3) ambiguous cases, where visual triage paths implied conditional logic not expressible as a single directed edge, were resolved by the expert and annotated with explanatory notes. This clinical authorship of the graph establishes validity at the source: all generated questions inherit their accuracy from expert-constructed relationships rather than requiring post-hoc review of generated outputs.

\subsection{Evaluation Query Instantiation}

We employ graph traversal to automatically instantiate MCQA evaluation queries that ensure complete coverage of medical relationships. For each condition node, we traverse its connected nodes to instantiate the five question types shown in Table~\ref{tab:question_examples}.

\begin{table*}[t]
\centering
\caption{Examples of auto-generated questions by relationship type.}
\label{tab:question_examples}
\small
\begin{tabular}{@{}lp{13cm}@{}}
\toprule
\textbf{Type} & \textbf{Example} \\
\midrule
Condition $\rightarrow$ Symptom &
\textbf{Q:} A 2 year old child with Very Severe Disease would most likely present with which symptom? \\
& \textbf{Options:} A: convulsions, B: chest indrawing, C: pus draining from the eye, D: WFH/L 2 z-scores or more\\
& \textbf{Answer:} A \\
\midrule
Symptom $\rightarrow$ Condition &
\textbf{Q:} A 21 month old child presenting with convulsions is most likely to have: \\
& \textbf{Options:} A: Cough or Cold, B: Very Severe Disease, C: Severe Pneumonia or Very Severe Disease, D: Very Severe Febrile Disease with no Malaria Risk\\
& \textbf{Answer:} B \\
\midrule
Condition $\rightarrow$ Treatment &
\textbf{Q:} Which treatment is recommended for a 21 month old child with Very Severe Disease? \\
& \textbf{Options:} A: assess or refer for TB assessment and INH preventive therapy, B: if mouth ulcers treat with gentian violet, C: do virological test at age 4--6 weeks or repeat 6 weeks after the child stops breastfeeding, D: give first dose of intramuscular antibiotics\\
& \textbf{Answer:} D \\
\midrule
Condition $\rightarrow$ FollowUp &
\textbf{Q:} What is the appropriate follow-up schedule for a 3 year old child with Some Dehydration? \\
& \textbf{Options:} A: follow-up in 14 days, B: follow-up in 5 days, C: follow-up in 2 days if not improving, D: follow-up in 7 days\\
& \textbf{Answer:} C \\
\midrule
Condition $\rightarrow$ Severity &
\textbf{Q:} A 13 month old child with Very Severe Disease should be classified as: \\
& \textbf{Options:} A: moderate, B: mild, C: none, D: severe\\
& \textbf{Answer:} D \\
\bottomrule
\end{tabular}
\end{table*}

The framework dynamically instantiates evaluation queries using four templates for each of five question types while maintaining clinical relevance and variability. Random age generation is constrained to the condition's valid range (e.g., 0--8 weeks for young infants, 2--60 months for children).

The distractor sampling algorithm prioritizes clinical validity through age-stratified selection. For each question requiring $k=3$ distractors, the system first identifies all conditions sharing the same age range as the target condition, creating an age-appropriate candidate pool.

For a question with correct answer $v_{\mathrm{corr}}$ of type $\tau$ and target condition with age range $\alpha$, we construct an age-appropriate distractor pool by selecting candidate nodes that (i) match the required type and (ii) are compatible with the target age range. Distractors are then sampled uniformly without replacement from this pool.

This construction ensures that all distractors are clinically plausible within the relevant age group while maintaining variability across generated questions. A formal specification of the distractor construction is provided in Appendix~\ref{app:distractor}.

The dynamic generation process creates novel evaluation instances through variation in templates, ages, and distractors while maintaining consistent difficulty and clinical relevance. This mitigates a key limitation of static benchmarks, in which models may have seen evaluation questions during training, while enabling substantial variation for robust statistical analysis.

\subsection{Contamination Resistance}
The harness addresses two distinct contamination risks that static benchmarks cannot mitigate.

\textbf{Surface-form contamination} occurs when evaluation questions appear verbatim in training data. By generating questions at evaluation time with randomized ages, distractor 
sampling, and template selection, the probability of repeated surface forms 
is reduced relative to static benchmarks; the valid combinatorial space is 
bounded in practice by clinical constraints on age--condition--distractor 
compatibility, as discussed in Section~\ref{sec:limitations}.

\textbf{Relationship-level contamination} occurs when a model has learned the underlying clinical relationships from source documents, such that it can answer questions correctly regardless of surface form. Unlike surface-form contamination, this cannot be mitigated through variation in phrasing alone.

Rather than attempting to eliminate this form of contamination, the proposed harness enables a complementary evaluation strategy. Because evaluation queries are generated dynamically from a structured representation of the guidelines, the same framework can be applied to updated or modified guidelines that postdate model training. This allows evaluation to probe whether models have genuinely acquired generalizable clinical reasoning or are relying on memorized relationships from specific guideline versions.

In this sense, the harness supports temporal and versioned evaluation, making it possible to identify knowledge gaps as clinical guidelines evolve. This shifts evaluation from static benchmarking to continuously refreshable assessment aligned with evolving domain knowledge.

Graph-level errors represent a third risk, where inaccuracies in the knowledge graph propagate to all generated questions. Expert authorship of the graph (Section~\ref{sec:expert}) directly addresses this by establishing the graph as a clinically verified source of evaluation truth.

\section{Case Study: WHO IMCI}

\subsection{Clinical Expert Authorship and Validation}
\label{sec:expert}

The knowledge graph underlying all generated questions was manually curated by a co-author who is a board-certified physician with over 15 years of clinical practice, specialized pediatric training, and extensive experience implementing WHO IMCI guidelines in clinical settings in sub-Saharan Africa. This authorship model, where domain expertise is embedded at the graph construction stage rather than applied as post-hoc review, provides stronger validity guarantees than question-level annotation alone: every generated question inherits its clinical accuracy from expert-constructed graph relationships.

To further validate the generated question set, the same expert reviewed the 432 auto-generated questions across the five relationship types: Condition $\rightarrow$ Treatment (130), Symptom $\rightarrow$ Condition (118), Condition $\rightarrow$ Symptom (118), Condition $\rightarrow$ Severity (37), and Condition $\rightarrow$ FollowUp (29). For each question, the review assessed: (1) clinical accuracy of the correct answer, (2) appropriateness of distractors for the specified age range, and (3) clarity and unambiguity of question phrasing. Given that questions are derived from an expert-curated graph, this review serves primarily to verify that the generation pipeline correctly traverses and formats the underlying relationships rather than to establish clinical accuracy de novo.

The graph was curated by a single clinical expert, which precludes inter-rater reliability assessment. The underlying guidelines provide deterministic decision rules, which partially mitigates subjectivity in annotation. Independent validation by additional clinicians with IMCI expertise remains important future work for establishing the rigor required of a production evaluation instrument.

\subsection{LLM Inference Results}

We conduct baseline inference evaluation to assess out-of-the-box model performance for the closed-source models Claude Sonnet 4.6, o4-mini, and GPT-5.2, the open-weights model GPT-OSS-20B, and the domain fine-tuned model MedGemma-4B. Models are compared across size and training regime, including closed-source frontier, open-weights, and domain fine-tuned, to characterize the performance 
landscape broadly; within-class comparisons are deferred to future work.  Models receive questions in a standardized format with explicit instructions to respond with only the letter (A, B, C, or D) corresponding to the correct answer. We measure accuracy per question type with uncertainty over the template variations.

\begin{figure*}[t]
\centering
\includegraphics[width=\textwidth]{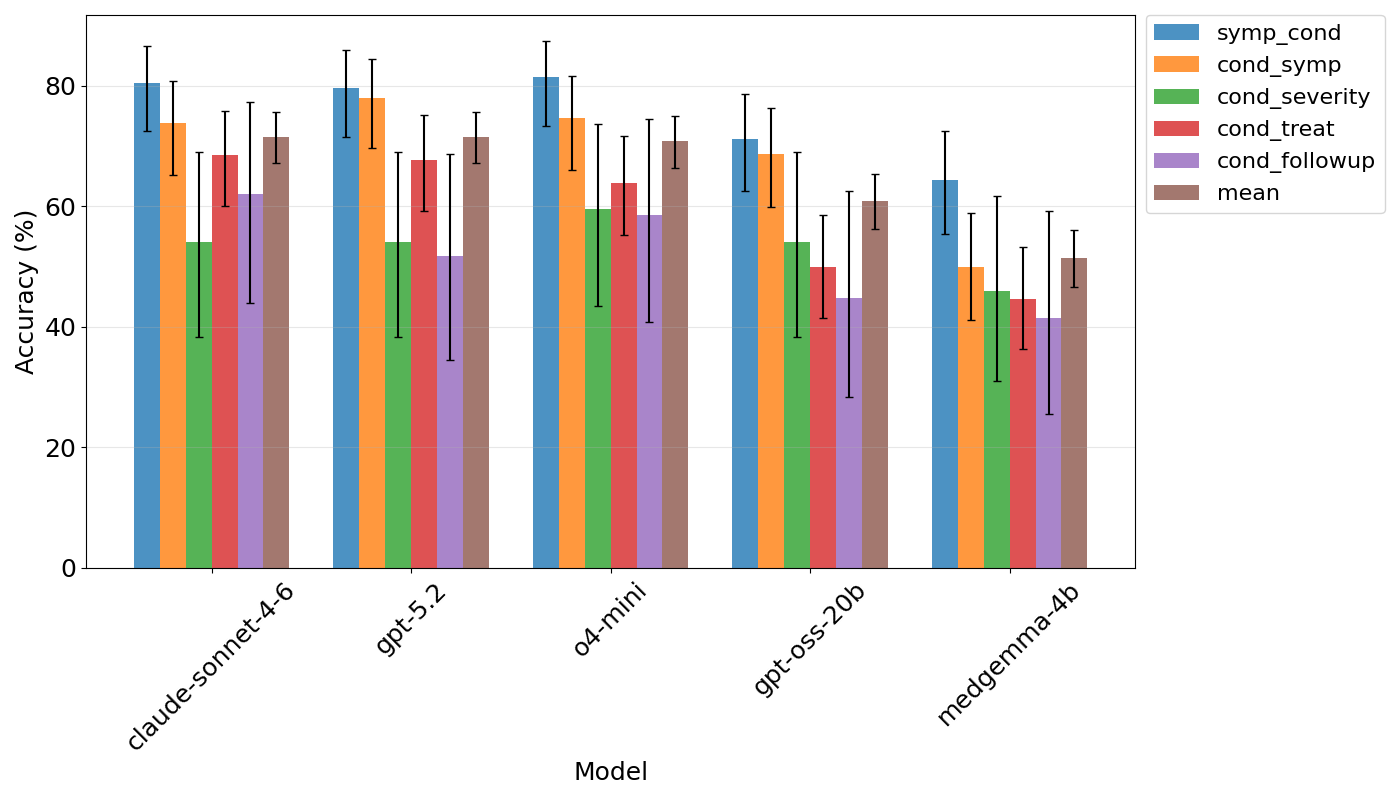}
\caption{Model accuracy across five clinical question categories: condition-symptom (C$\rightarrow$S), symptom-condition (S$\rightarrow$C), condition-treatment (C$\rightarrow$T), condition-severity (C$\rightarrow$Sv), and condition-followup (C$\rightarrow$F), along with overall mean accuracy across all categories. Error bars represent 95\% Wilson score confidence intervals computed at the question level, treating each question as an independent Bernoulli trial.}
\label{fig:accuracy_all_types_ci}
\end{figure*}

Figure~\ref{fig:accuracy_all_types_ci} and Table~\ref{tab:model_performance} present model performance across question types.

\begin{table*}[t]
\centering
\caption{Model accuracy (\%) on the IMCI knowledge graph evaluation across five clinical question categories: condition-symptom (C$\rightarrow$S), symptom-condition (S$\rightarrow$C), condition-treatment (C$\rightarrow$T), condition-severity (C$\rightarrow$Sv), and condition-followup (C$\rightarrow$F). Values are reported as accuracy $\pm$ the half-width of the 95\% Wilson score confidence interval, computed at the question level. Overall accuracy is pooled across all questions (question-weighted). Bold indicates the highest accuracy in each column.}
\label{tab:model_performance}
\footnotesize
\begin{tabular}{@{}lcccccc@{}}
\toprule
\textbf{Model} & \textbf{Overall} & \textbf{C$\rightarrow$S} & \textbf{S$\rightarrow$C} & \textbf{C$\rightarrow$T} & \textbf{C$\rightarrow$Sv} & \textbf{C$\rightarrow$F} \\
\midrule
Claude Sonnet 4.6 & \textbf{72.0$\pm$4.2} & 73.7$\pm$7.8 & 80.5$\pm$7.1 & \textbf{68.5$\pm$7.9} & 54.0$\pm$15.3 & \textbf{62.1$\pm$16.6} \\
GPT-5.2           & \textbf{72.0$\pm$4.2} & \textbf{78.0$\pm$7.4} & 79.7$\pm$7.2 & 67.7$\pm$7.9 & 54.0$\pm$15.3 & 51.7$\pm$17.1 \\
o4-mini           & 71.0$\pm$4.3 & 74.6$\pm$7.8 & \textbf{81.4$\pm$7.0} & 63.9$\pm$8.2 & \textbf{59.5$\pm$15.1} & 58.6$\pm$16.9 \\
GPT-OSS-20B       & 61.0$\pm$4.6 & 68.6$\pm$8.3 & 71.2$\pm$8.1 & 50.0$\pm$8.5 & 54.0$\pm$15.3 & 44.8$\pm$17.0 \\
MedGemma-4B       & 51.0$\pm$4.7 & 50.0$\pm$8.9 & 64.4$\pm$8.5 & 44.6$\pm$8.4 & 46.0$\pm$15.3 & 41.4$\pm$16.9 \\
\bottomrule
\end{tabular}
\end{table*}

\begin{figure*}[t]
\centering
\includegraphics[width=\textwidth]{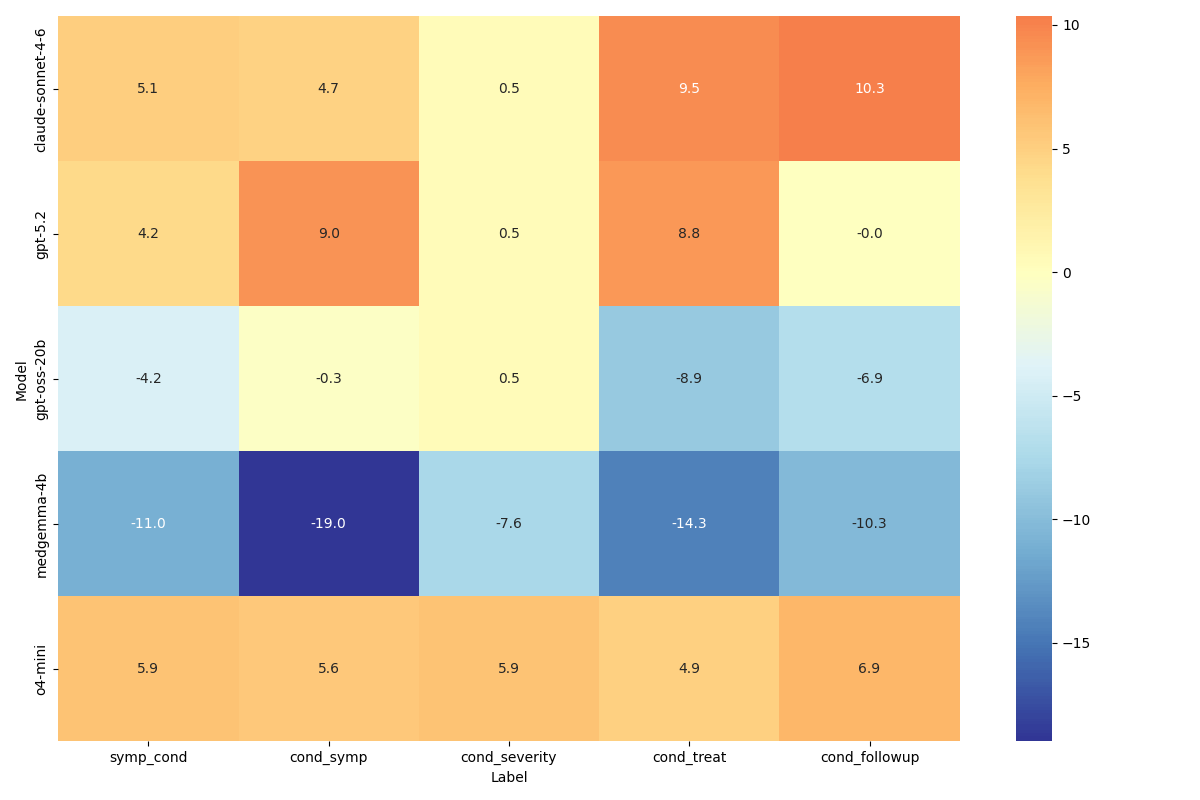}
\caption{Accuracy delta heatmap showing the difference between question-type-specific accuracy and overall model accuracy for each model. Positive values (red/orange) indicate above-average performance for that question type, while negative values (blue) indicate below-average performance. Values are expressed as percentage points.}
\label{fig:accuracy_delta_heatmap}
\end{figure*}

Figure~\ref{fig:accuracy_delta_heatmap} presents model performance variations across clinical question types, measured as the delta between question-specific accuracy and overall model accuracy.

\subsection{Key Findings}

\begin{enumerate}
    \item The three frontier closed-source models, Claude Sonnet 4.6, GPT-5.2, and o4-mini, achieve statistically indistinguishable overall accuracy (71--72\%), as their confidence intervals overlap substantially. The smaller models GPT-OSS-20B (61\%) and MedGemma-4B (51\%) perform well above random (25\%).
    \item Symptom $\rightarrow$ Condition questions show the highest performance across all models (64--81\%), indicating that models better recognize symptoms than prescribe treatments or protocols.
    \item Within-model performance varies substantially across question types, underscoring that aggregate accuracy obscures meaningful capability differences.
    \item MedGemma-4B has lower performance than larger models across all question types, indicating that model scale and general reasoning capacity may dominate performance in this setting.
\end{enumerate}

Unlike human-curated benchmarks, our dynamic graph-based method ensures complete coverage of all guideline relationships, consistent terminology from source documents, reduced data contamination through automated generation, and scalability to other medical guidelines.

\subsection{Template Ablation Study}

\begin{figure*}[t]
\centering
\includegraphics[width=\textwidth]{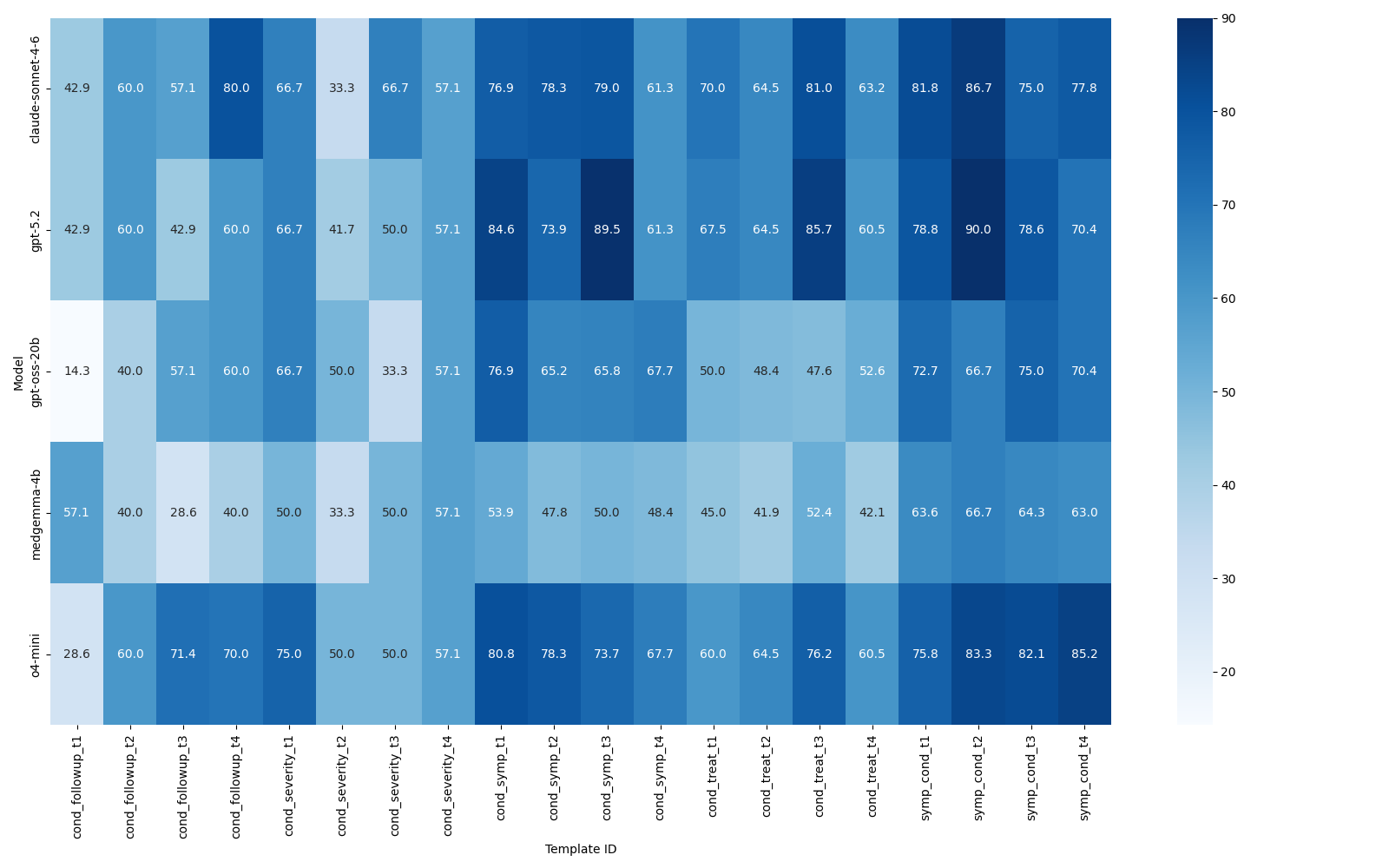}
\caption{Accuracy by template and model. Each cell shows the accuracy (\%) for a given model and question template. Darker blue indicates higher accuracy. Substantial within-type variance across templates demonstrates that question phrasing affects model performance independently of the underlying clinical relationship.}
\label{fig:template_heatmap}
\end{figure*}

Figure~\ref{fig:template_heatmap} reveals substantial within-type variance across question templates, demonstrating that phrasing significantly affects model performance independently of the underlying clinical relationship being tested. The \texttt{cond\_followup\_t1} template (``When should a \{age\} old child with \{cond\} return for follow-up?'') consistently produces the lowest accuracy across all models (14--57\%), while \texttt{cond\_symp\_t3} produces some of the highest (50--90\%). This variance has direct implications for evaluation harness design: using multiple templates per question type, as our harness does, provides more robust estimates of model capability than single-template approaches, and averaging over template variants reduces the influence of phrasing artifacts on reported accuracy.

\section{Evaluation Considerations}
\label{sec:infrastructure}

\subsection{Operationalization}
A key question for operationalization is how performance on this benchmark 
translates to real-world deployment. We argue that grounding evaluation in 
WHO guidelines provides a meaningful bridge: because the guidelines represent 
human-reviewed, authoritative clinical decision logic, high performance on 
graph-derived questions indicates alignment with expert-validated protocols. 
This supports a unit and integration testing analogy: unit tests verify that 
a model correctly handles individual clinical relationships (e.g., symptom 
$\rightarrow$ condition), while integration tests verify coherent reasoning 
across chains of relationships (e.g., symptom $\rightarrow$ condition 
$\rightarrow$ treatment $\rightarrow$ follow-up). While MCQA cannot capture 
the full complexity of patient vignettes or multi-turn clinical conversations, 
its discrete, unambiguous structure makes it well-suited for unit testing: 
each question has a single correct answer that requires no rubric to grade. 
The dynamic nature of the harness further strengthens this analogy, because 
questions are instantiated at evaluation time from the graph rather than drawn 
from a fixed set, models cannot memorize the test suite, preserving the 
integrity of repeated evaluation as guidelines and models evolve. In practice, 
this enables two concrete deployment decisions: models that fall below 
acceptable performance thresholds on clinically critical question types can 
be replaced by better-performing alternatives, and if a frontier model's 
guardrails change, a known risk in health domains where medically valid 
questions can trigger content filters, the harness provides a reproducible 
basis for selecting a replacement model with documented clinical protocol 
alignment.

\subsection{Cost and Scalability Relative to Manual Curation}

Manual benchmark curation requires domain experts to author, review, and validate each question individually, a process that does not scale and produces static artifacts vulnerable to contamination. Our harness shifts the labor from question authorship to graph construction: a one-time cost that yields a large and refreshable space of evaluation instances for practical evaluation. For IMCI, manual curation of the knowledge graph by a domain clinical expert required significant upfront investment, after which 432 questions were generated automatically with validity inherited from the graph structure. Expanding across the combinatorial space induced by templates, ages, and distractors requires no additional expert labor beyond graph maintenance as guidelines are updated.

The primary scaling bottleneck is graph construction itself. Automated extraction via PDF parsers and LLMs missed critical relationships because the conditional logic in IMCI flowcharts is expressed visually through color-coded triage paths and nested decision branches that current pipelines cannot faithfully reconstruct as directed edges. Future work could reduce this bottleneck through semi-automated graph construction with expert review, particularly for guidelines with consistent structure such as WHO protocols.

\subsection{Stakeholder Roles}
The harness separates evaluation into three distinct stakeholder roles with different expertise requirements. \textit{Graph constructors} require deep domain expertise to accurately encode guideline relationships; in our case, a pediatrician with IMCI implementation experience in sub-Saharan Africa. \textit{Harness operators} require technical expertise to run generation and evaluation pipelines but not medical knowledge. \textit{Model developers} can consume evaluation results without access to the underlying graph, enabling third-party evaluation with separation between evaluators and developers, a property the EvalEval community has identified as important for accountability \cite{reuel2025whoevaluates}.

This separation also clarifies accountability: errors in evaluation results can be traced to graph inaccuracies (domain expert responsibility), generation bugs (harness operator responsibility), or model failures (developer responsibility).

\subsection{Extensibility to Other Guidelines}
The graph schema, including conditions, symptoms, treatments, follow-ups, severities, and their directed relationships, is not specific to IMCI. Any clinical guideline with structured decision logic is a candidate. WHO produces guidelines across malaria, tuberculosis, HIV, and maternal health that share the same flowchart structure as IMCI. Beyond healthcare, structured regulatory guidelines, legal compliance frameworks, and technical standards with explicit relationship structures could support the same approach. The primary requirement is that the source document encodes relationships explicitly enough to support graph construction, a property common to clinical and regulatory guidelines by design.

\subsection{Limitations}
\label{sec:limitations}
Question quality depends entirely on graph accuracy: any errors in manual 
annotation propagate to all generated questions. The graph was curated by 
a single clinical expert, which precludes inter-rater reliability assessment; 
independent validation by additional clinicians with IMCI expertise remains 
important future work for establishing the rigor required of a production 
evaluation instrument. We evaluate only MCQA format, which cannot capture 
the complexity of real clinical reasoning involving differential diagnosis 
and incomplete information. Our text-only approach excludes visual diagnostic 
elements present in the original IMCI handbook. While question generation is 
automated, initial graph construction remains manual, limiting scalability. 
Our evaluation on IMCI guidelines may not generalize to other medical domains. 
Although the framework admits a large combinatorial space of possible instances, 
the practically valid subset is smaller because clinical constraints introduce 
dependencies among age, condition, and distractor choices, and we have not 
exhaustively verified all such variants. Finally, the absence of a human expert 
baseline makes it difficult to interpret absolute model accuracy; frontier models 
scoring 71--72\% may represent strong or weak performance depending on task 
difficulty, and establishing a human ceiling is an important direction for 
future work.

\subsection{Potential Risks}
This work presents evaluation tools for medical AI systems. Models performing well on MCQA may still fail in actual clinical scenarios requiring differential diagnosis and incomplete information. Any errors in manual graph annotation propagate to evaluation, potentially validating incorrect medical knowledge. Our focus on WHO IMCI guidelines may not generalize to other healthcare contexts. This evaluation harness is intended for research purposes only and is not suitable for clinical decision-making.

\section{Conclusion}

This work introduces a graph-based evaluation harness for systematically instantiating evaluation queries from clinical guidelines, demonstrated on the WHO IMCI handbook. By transforming medical guidelines into queryable graphs, the framework achieves complete coverage of encoded relationships, which is not feasible through manual curation alone. Its dynamic design allows new evaluation instances with different ages and distractors to be sampled continuously, including as guidelines are updated. While baseline inference provides initial scores, the main value lies in granular performance across relationship types, which reveals systematic strengths and weaknesses in clinical protocol understanding.

The clinical validity of the generated questions rests on expert authorship of the underlying graph rather than post-hoc sampling, a design choice that both strengthens the validity claim and clarifies the role of domain expertise in evaluation infrastructure. The graph-based approach is extensible beyond IMCI, addressing the gap between general-purpose benchmarks and real-world domain-specific applications.

\bibliography{main}

\appendix

\section{Distractor Pool Construction}
\label{app:distractor}
We formalize distractor construction for completeness. Let $G=(V,E)$ denote the IMCI knowledge graph.

For a question with correct answer $v_{\mathrm{corr}}$ of type $\tau$ and age
range $\alpha$, the distractor pool is defined as

\begin{equation}
P_{\tau,\alpha} =
\begin{cases}
C_\alpha \setminus \{v_{\mathrm{corr}}\}, & \tau = \mathrm{Cond}, \\
\mathcal{N}_{\tau,\alpha} \setminus \{v_{\mathrm{corr}}\},
& \tau \in \mathcal{T}, \\
S \setminus \{v_{\mathrm{corr}}\}, & \tau = \mathrm{Sev}.
\end{cases}
\end{equation}

where $\mathcal{T}=\{\mathrm{Sym},\mathrm{Treat},\mathrm{FollowUp}\}$.

The condition set is
\begin{equation}
C_\alpha = \{c \in V : \mathrm{type}(c)=\mathrm{Condition},\
\mathrm{age\_range}(c)=\alpha\},
\end{equation}

and the aggregated neighborhood is
\begin{equation}
\mathcal{N}_{\tau,\alpha} = \bigcup_{c \in C_\alpha} N_\tau(c).
\end{equation}

The neighborhood function is
\begin{equation}
N_\tau(c)=
\begin{cases}
\{u : (u,c)\in E,\ \mathrm{type}(u)=\tau\}, & \tau=\mathrm{Sym}, \\
\{u : (c,u)\in E,\ \mathrm{type}(u)=\tau\},
& \tau \in \{\mathrm{Treat},\mathrm{FollowUp}\}.
\end{cases}
\end{equation}

For severity classification,
\begin{equation}
S = \{u \in V : \mathrm{type}(u)=\mathrm{Severity}\}.
\end{equation}

The final distractor set is
\begin{equation}
D = \mathrm{sample}(P_{\tau,\alpha}, k),
\end{equation}
where $k=3$.

\end{document}